\documentclass{article}

\usepackage[nonatbib,final]{nips_2017}
\usepackage[utf8]{inputenc} 
\usepackage[T1]{fontenc}    
\usepackage{hyperref}       
\usepackage{url}            
\usepackage{booktabs}       
\usepackage{amsfonts}       
\usepackage{nicefrac}       
\usepackage{microtype}      

\usepackage[linesnumbered,ruled]{algorithm2e}
\usepackage{color}
\usepackage{graphicx}
\usepackage{amsmath}
\usepackage{mathtools}
\title{
Sum of previous inpatient serum creatinine measurements; a practical model for acute kidney injury in rehospitalized patients
}

\author{
  Samuel J. Weisenthal$^{1,2}$, 
  Haofu Liao$^{3}$,
    Philip Ng$^{4}$,
    and Martin S. Zand$^{1,2,5}$\\
 \small
  1. University of Rochester School of Medicine and Dentistry, University of Rochester Medical Center \\
  \small
  2. Rochester Center for Health Informatics \\
 \small
  3. Department of Computer Science, University of Rochester \\
    \small
  4. University of Rochester Information Systems Division\\ 
  \small
  5. Department of Medicine, Division of Nephrology \\ 
}

\begin{document}

\maketitle

\begin{abstract}
Acute Kidney Injury (AKI), the abrupt decline in kidney function due to temporary or permanent injury, is associated with increased mortality, morbidity, length of stay, and hospital cost.  Sometimes, simple interventions such as medication review or hydration can prevent AKI.  There is therefore interest in estimating risk of AKI at hospitalization.  
To gain insight into this task, we employ multilayer perceptron (MLP) and recurrent neural networks (RNNs) using serum creatinine (sCr) as a lone feature.  We explore different feature input structures, including variable-length look-backs and a nested formulation for rehospitalized patients with previous sCr measurements.  
Experimental results show that the simplest model, MLP processing the sum of sCr, had best performance: AUROC 0.92 and AUPRC 0.70.  Such a simple model could be easily integrated into an EHR.  Preliminary results also suggest that inpatient data streams with missing outpatient measurements---common in the medical setting---might be best modeled with a tailored architecture. 
\end{abstract}

\section{Introduction}


Acute Kidney Injury (AKI), the abrupt decline in kidney function due to temporary or permanent injury, is associated with increased mortality, morbidity, length of stay, and hospital cost~\cite{chertow2005MortLOSCost}.  There exist a variety of preventative strategies~\cite{lameire2008prevention}---some types of AKI (e.g., from non-steroidal anti-inflammatory medication, radiocontrast chemotherapy, or aminoglycoside antibiotics) can be prevented outright by altering treatment or close monitoring.
For this reason, there is particular interest in modeling AKI with electronic health record (EHR) data~\cite{sutherland2016utilizingAKIADQI}. 
Prior hospitalizations generate enormous amounts of data, some of it inaccessible to clinicians via current interfaces; we hope to gain insight into the way this data might be leveraged to help predict and prevent AKI.  Of particular interest as a predictor is serum creatinine (sCr). Creatinine is a protein that accumulates in the serum if kidney filtration is reduced, acting as a surrogate for the true measure of kidney function, glomerular filtration rate (GFR).  In this study we process sCr with recurrent neural networks (RNNs)~\cite{rumelhart1986sequential} and multilayer perceptrons (MLPs) using different input structures.  

\section{Related work and background}
\begin{figure}[!ht]
\centering
\includegraphics[width=0.7\linewidth]{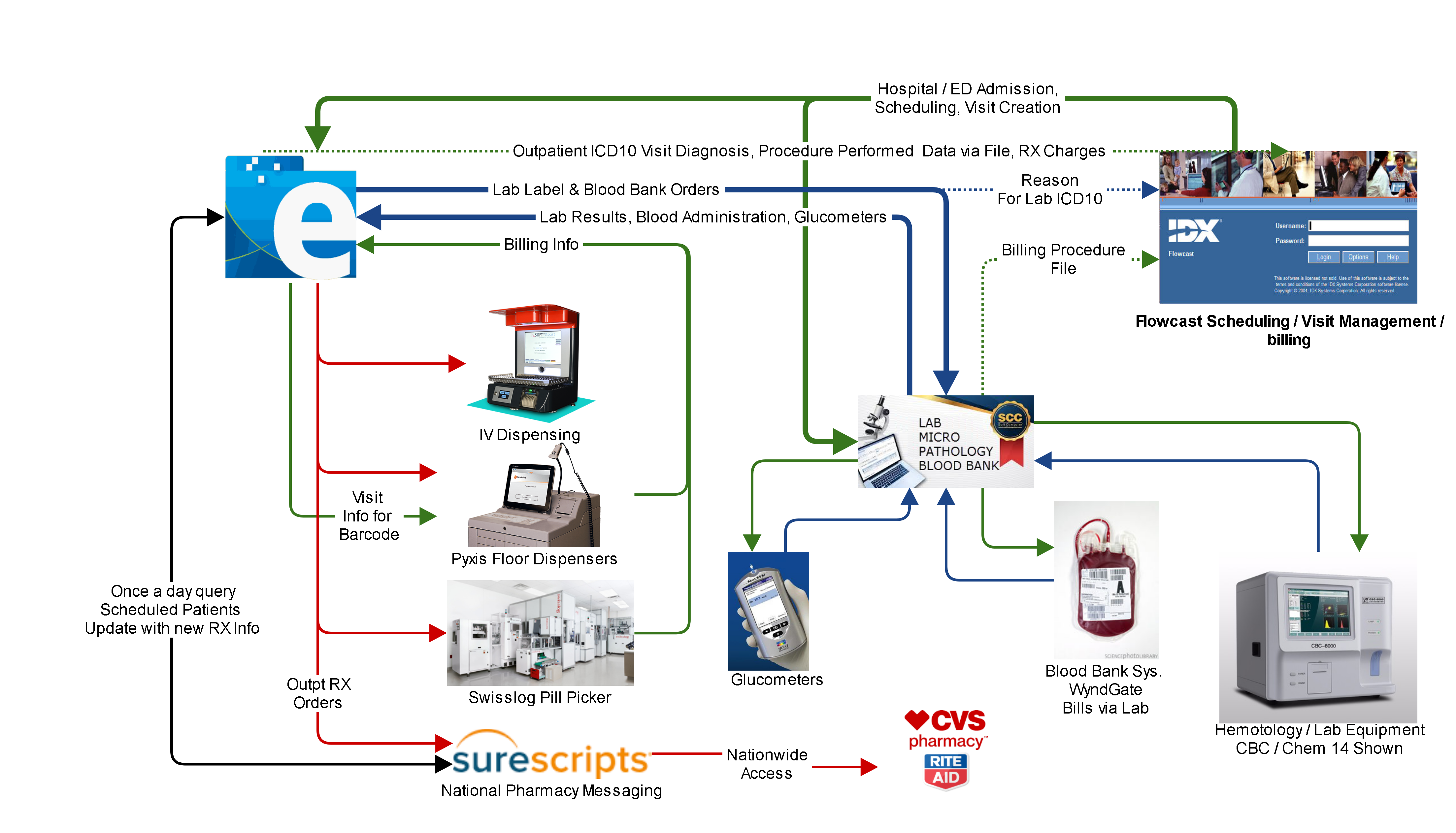}
\caption{{\bf EHR diagram.} Blue lines are lab data flow, green billing and scheduling, and red medications.  Dotted lines correspond to data files and solid to direct Health Level 7 integration.}
\label{EHR_fig}
\end{figure}

AKI prediction is an active area of research, with special emphasis on features from the
the Electronic Health Record (EHR) data~\cite{koyner2016development,sutherland2016utilizingAKIADQI}.  
In particular, there is interest in 
construction of models that apply to a broad patient population~\cite{sutherland2016utilizingAKIADQI}.
Many current models focus on AKI in the context of cardiac procedures,
the critically ill,
the elderly,
liver
and lung
transplant patients, and rhabdomyolsis.
Most
use logistic regression, and although 
some use decision trees 
or ensemble methods. 
These models use features from the current hospitalization; in contrast, in line with \cite{choi2016doctor}, we use features from previous visits to estimate the probability of AKI in a rehospitalization given data from prior hospitalizations, focusing on the cohort of patients who are rehospitalized and also have previous sCr measurements.  This particular cohort has high prior probability of AKI and therefore a predictive model could have real application; e.g., many rehospitalized patients present with conditions that may benefit from certain medications best administered when AKI risk is low.  By considering only longitudinal sCr data, we also explore a much simpler, more interpretable model space than other studies.  The EHR data flow is complex (Figure~\ref{EHR_fig}) and implementing models that depend on many features might be difficult, the models described here might more easily be implemented into an EHR, facilitating translation into the clinic.

RNNs are popular in medical research; they are used for phenotyping in~\cite{lipton2015learning} and heart failure detection and next visit prediction in~\cite{choi2016doctor,choi2016HF}. 
In~\cite{choi2016doctor}, the authors predicted AKI among other diagnosis codes, inspiring us to pursue it further with laboratory values in the inpatient setting.  In~\cite{choi2016doctor}, features are collected per unit of time, whereas here we only consider the order of measurements, but not their actual time stamps.
Since we process sequences of measurements, there is related work in natural language modeling.  
As phrases can be represented by averaging words~\cite{mikolov2013efficient}, hospitalizations might be represented by averaging measurements.  
Hierarchical RNN architectures have been employed ~\cite{sordoni2015hierarchical,chung2016hierarchical} for context awareness and boundary inference.

\section{Methods}
We use an IRB-approved, de-identified adult dataset from an inpatient EHR system.  The full dataset contains roughly six million laboratory records. We assigned a positive label if AKI was coded (according to International Statistical Classification of Diseases ICD $9^{th}$ edition) or sCr trajectories met the most current diagnostic criteria~\cite{ERBPpositionState}. 
The data for each patient consists of a variable-length sequence of hospitalizations, or hospitalizations, where each hospitalization is a variable-length sequence of measurements. We have no information on these patients outside of their hospitalizations.
 We restrict ourselves to a single feature, sCr
to build a simple, interpretable model.
We forward fill missing sCrs (since we processed sequences, there were very few of these, so we did not use a more sophisticated method; these were truly missing at random because a test was recorded, but no result value was present).  We denote sCr measurement from hospitalization $a$ at time $t$ as $m_{(a)}^{(t)}$.  Each patient could therefore be represented as a list of $A$ hospitalizations, each of which contained a list of $\tau_{(a)}$ sCr measurements. 
$$[
1:[m_{(1)}^{(1)}, ... ,m_{(1)}^{(\tau_{(1)})}],\\
... ,\\
A:[m_{(A)}^{(1)}, ... ,m_{(A)}^{(\tau_{(A)})}]$$
There are $A$ labels, where each is an indicator of whether AKI occurred in the hospitalization directly following hospitalization $a$.
$$[l_{1}, ... ,l_{A}]$$

\subsection{Input feature structures}
The basic formulations are as follows (further details are provided in the Appendix): each hospitalization-label pair can be considered independently (MARKOV)
\footnote{We denote this one as MARKOV because it is memoryless past one hospitalization}; all prior measurements can be concatenated (CONCAT); we can consider all observations from the same patient as a single, nested sequence (NEST).

For processing by MLP, the measurements must be aggregated. We can choose MARKOV or CONCAT before aggregating and then employ MEAN, SUM, or MAX.  Data processing was performed in Pandas~\cite{mckinney2010data}.


\subsection{Architectures}
We do not review the standard RNN architecture in entirety, but we try to follow the notation in Goodfellow et al.~\cite{GoodFellowBengio2015deep}, which provides more detail.  For more detail on our methods, see the Appendix.  We use cross-entropy loss and hyperbolic tangent activation in all experiments.  We use a many-to-one RNN 
for MARKOV and CONCAT.   

For the NEST input structure we require an RNN that processes nested, variable-length sequences where each inner sequence produces only a single output. We therefore modified the RNN architecture to process nested sequences by chaining together multiple instances of the many-to-one network.  We suspected that information might be passed differently from one hospitalization to the next than it is from one measurement to the next, so we introduced a new "rehospitalization" parameter, $R$, that was only between hospitalizations.  Since our data consists of hospitalization sequences that contain measurement sequences, we can index the new equations by $1\leq a \leq A$ where each $a$ has $\tau_{(a)}$ measurements.  With $R$, we have the forward equations

 

$$
z^{(t)}_{(a)} = \left\{
        \begin{array}{ll}
            Wm^{(t)} _{(a)}+Rh^{(\tau _ {(a-1)})}_{(a-1)}+r  & \quad t= 1 \\
            
  Wm^{(t)} _{(a)}+Uh^{(t-1)}_{(a)}+b  & \quad t \neq 1 \\
        \end{array}
    \right.
$$

$$h^{(t)} _{(a)}=\tanh(z^{(t)} _{(a)})$$
$$y _{(a)}= Vh^{(\tau _ {(a)})} _{(a)} + c$$



 
  
 


The network is now unrolled in time over measurements and over hospitalizations before backpropagation.  All models were implemented using Numpy~\cite{walt2011numpy}.

\section{Experiments}
The dataset contains 135,862 sCr measurements from 12,491 unique patients who generated 26,606 hospitalizations.  AKI occurred in rehospitalization after 15.7\% of the hospitalizations. Throughout the dataset, there were on average 2.1 $\pm$ 2.4 hospitalizations per patient and on average 5.1 $\pm$ 9.9 sCr measurements per hospitalization. Over 100 trials, we compare the three input structures MARKOV, CONCAT, and NEST where NEST has the inter-hospitalization variable R (otherwise, it is identical to CONCAT).  We also evaluate an MLP acting on MARKOV or CONCAT with aggregation function SUM, MEAN, or MAX.  Each trial has identical (random) parameter initializations and the same shuffled training/validation dataset.  Note that since the MLP has no parameter $U$ linking hidden states, we were not able to initialize it with the same parameters as the RNNs, but inter-MLP comparisons have identical parameter initializations as well. We explore different numbers of hidden units (HUs) (10, 50, and 100).
Twenty percent of the full dataset was held out as test data and 20\% of the training data was held out as validation data (the prevalence of AKI upon rehospitalization in the train and test sets were roughly equal at 16\% and 15\%, respectively).  (20\% is still held out.)  For all splits, we ensure that no hospitalizations from the same patient are in the training and testing set.  We trained each model for 20 epochs using AdaGrad~\cite{duchi2011adaptive} and then tested the model with the best validation set AUROC of the 100 trials on the test set.  
The distribution of area under the receiver operating characteristic (AUROC), area under the precision recall curve (AUPRC), and logistic loss (LL) for the validation set over the 100 trials with different numbers of hidden units are shown in the Appendix (Figure \ref{fig3}). We report untouched test set performance for the best models in Table \ref{tr}.  All metrics are computed at the hospitalization (not patient) level, and therefore patients with multiple hospitalizations are represented multiple times.  Evaluation metrics were computed with functions from scikit-learn~\cite{pedregosa2011scikit}.

\begin{table}[ht]
  \caption{Held-out test set performance}
  \label{tr}
  \tiny 
  \centering

\begin{tabular}{lllrrr}
\toprule
\# HU & Model & Input Struct & LL &     AUPRC &     AUROC \\
\midrule
10 & RNN & NEST &  0.438376 &  0.460836 &  0.831527 \\
   &     & MARKOV &  0.422027 &  0.614550 &  0.900993 \\
   &     & CONCAT &  0.367677 &  0.460945 &  0.831687 \\
   \cmidrule{3-6}
   & MLP & MARKOV-MAX &  0.373659 &  0.507971 &  0.843675 \\
   &     & CONCAT-MAX &  0.381322 &  0.470231 &  0.832244 \\
   &     & MARKOV-MEAN &  0.391130 &  0.416539 &  0.765329 \\
   &     & CONCAT-MEAN &  0.392883 &  0.402276 &  0.758172 \\
   &     & MARKOV-SUM &  0.281967 &  \textbf{0.697588} &  \textbf{0.919228} \\
   &     & CONCAT-SUM &  0.394754 &  0.479620 &  0.844405 \\
   \cmidrule{2-6}
50 & RNN & NEST &  0.475782 &  0.430490 &  0.786414 \\
   &     & MARKOV &  0.371971 &  0.560443 &  0.878528 \\
   &     & CONCAT &  0.421936 &  0.213510 &  0.690209 \\
   \cmidrule{3-6}
   & MLP & MARKOV-MAX &  0.375280 &  0.507971 &  0.843675 \\
   &     & CONCAT-MAX &  0.383305 &  0.470231 &  0.832244 \\
   &     & MARKOV-MEAN &  0.392830 &  0.416530 &  0.765314 \\
   &     & CONCAT-MEAN &  0.394410 &  0.402274 &  0.758167 \\
   &     & MARKOV-SUM &  \textbf{0.281964} &  0.697587 &  0.919227 \\
   &     & CONCAT-SUM &  0.374631 &  0.479623 &  0.844408 \\
   \cmidrule{2-6}
100 & RNN & NEST &  0.495673 &  0.471420 &  0.816330 \\
   &     & MARKOV &  0.425833 &  0.596208 &  0.879722 \\
   &     & CONCAT &  0.390785 &  0.211309 &  0.679677 \\
   \cmidrule{3-6}
   & MLP & MARKOV-MAX &  0.373959 &  0.507971 &  0.843675 \\
   &     & CONCAT-MAX &  0.380864 &  0.470231 &  0.832244 \\
   &     & MARKOV-MEAN &  0.392619 &  0.416536 &  0.765317 \\
   &     & CONCAT-MEAN &  0.393588 &  0.402281 &  0.758176 \\
   &     & MARKOV-SUM &  0.282694 &  0.697584 &  0.919224 \\
   &     & CONCAT-SUM &  0.387263 &  0.479623 &  0.844405 \\
\bottomrule
\end{tabular}

HU = hidden units; LL = log loss; AUPRC = area under PR curve; AUROC = area under ROC curve

\end{table}

\section{Conclusion}
Using only sCr, RNNs and MLP predict AKI in rehospitalizations with high AUROC and AUPRC.  
The best performing model was very simple.  MARKOV, where we treated each hospitalization as an independent sequence, was the best performing input structure for the RNN; MARKOV-SUM, where we treated each hospitalization as an independent sequence and aggregated it using sum, gave best results for the MLP and in general.  
Although this is still preliminary work and requires validation at different institutions, this is an exceedingly simple model and might be easily integrated into an EHR (since performance was insensitive to HU, a lower-capacity model like logistic regression could be appropriate).  Further, it should be noted that this model was actually a baseline and therefore its discovery was happenstance---it was not the original intent of the authors to evaluate it; a new study intended to externally validate the model would help ensure that the finding was not unique to this particular dataset.  Such a model is enticing however because the sum of sCr takes into account both sCr values---reflecting renal function, frequency of the orders--- reflecting physician concern, and length of the hospitalization (more work should be done to disentangle these factors, but capturing them all in a single input allows us to only estimate one parameter).

Our study also sheds light on how to model inpatient data when outpatient measurements are unknown. This is a common occurrence in medical datasets, many of which are generated by a single hospital lacking complete outpatient data.  For 10 HU, inserting a different parameter R between hospitalizations did not seem to affect results.  For greater than 10 HU, inserting R appeared to improve AUROC and AUPRC considerably but worsen LL. LL might have decreased because AUROC, not LL, was optimized for in model selection and the selected model was not required to be properly calibrated, but further experiments are needed to more fully explore this phenomenon.
These preliminary results however might suggest that a similar architecture or hierarchical RNN might be best for data streams with missing outpatient measurements---information flows differently between measurements than between hospitalizations.  

We stress our general finding of the benefits of investigating different permutations of input data structure (varying time window and aggregation function) and models for medical time-series prediction tasks.  The simple models yielded from this process might be very interpretable (simply summing all previous sCr measurements is easy to explain and could even be performed manually) and easy to implement into an EHR, which could allow quick adoption and facilitate the transition to automated "data-driven" healthcare practice, making way for more sophisticated techniques in the future.  Future directions include incorporating time stamps because more recent measurements might be most important and employing more sophisticated RNN architectures. 

\subsubsection*{Acknowledgments}

The project described in this publication was supported by the University of Rochester CTSA award number TL1 TR 002000 from the National Center for Advancing Translational Sciences of the National Institutes of Health. The content is solely the responsibility of the authors and does not necessarily represent the official views of the National Institutes of Health.  The authors acknowledge Robert White for obtaining the data and the anonymous reviewers for help in improving the manuscript.

\subsection*{Appendix}
\subsubsection*{Input structures}
\label{Appendix1}

\begin{itemize}
\item Each hospitalization-label pair can be considered independently (MARKOV).

$[
([m_{(1)}^{(1)}, ... ,m_{(1)}^{(\tau_{(1)})}],[l_{1}]), ... , ([m_{(A)}^{(1)}, ... ,m_{(A)}^{(\tau_{(A)})}],[l_{A}])]$

This corresponds to the time window approach with lag 1 visit in~\cite{choi2015doctor}. 
This kind of Markovian approach ignores measurements from all but the most recent hospitalization, and also does not take into account that hospitalizations are from the same patient.  Each sequence of measurements, however, is relatively short, alleviating concerns about vanishing or exploding gradients~\cite{bengio1994learning,hochreiter2001gradient}.
\item All prior measurements can be concatenated (CONCAT).

$[
([m_{(1)}^{(1)}, ... ,m_{(1)}^{(\tau_{(1)})}],[l_{1}]),... , ([m_{(1)}^{(1)}, ... ,m_{(1)}^{(\tau_{(1)})},...,m_{(A)}^{(1)}, ... ,m_{(A)}^{(\tau_{(A)})}],[l_{A}])
]$

Unlike MARKOV, CONCAT has memory for previous hospitalization measurements, but it still ignores that these hospitalizations are from the same patient. 

\item We can consider all observations from the same patient as a single, nested sequence (NEST).

$(
[
[
m_{(1)}^{(1)}, ... ,m_{(1)}^{(\tau_{(1)})}],
... ,
[m_{(A)}^{(1)}, ... ,m_{(A)}^{(\tau_{(A)})}]
]
,
[
l_{1},...,l_{A}]
)$

NEST should theoretically be preferable to MARKOV and CONCAT as all measurement information for each hospitalization is retained and also we take into account that the hospitalizations are from the same patient.  An RNN processing NEST requires nBPTT for training.  NEST produces a single sequence of the same length as the longest sequence in CONCAT.
\end{itemize}
For processing by MLP, the measurements must be aggregated. We can choose MARKOV or CONCAT before aggregating and then employ MEAN, SUM, or MAX.  

For example, to aggregate a MARKOV hospitalization using SUM, we just convert:\\
$$[
m_{(j)}^{(1)}, ... ,m_{(j)}^{(\tau_{(1)})}] \rightarrow
\sum_{i=1}^{\tau_{(1)}} m_{(j)}^{(i)}$$


Benefits of the aggregation approach are that it is parsimonious and SUM elegantly represents the number of measurements (weighted by magnitude)---this could be highly predictive since the frequency of medical testing often reflects a clinician's anxiety over a patient's status, which might be a strong indicator of deterioration, but the value of the test is also relevant.

\subsubsection*{RNNs}
\label{Appendix2}
Again we follow the notation in Goodfellow et al.~\cite{GoodFellowBengio2015deep}.  We only process a single scalar sCr, but provide the matrix equations in case of multi-dimensional input.  Each sCr measurement is denoted $m$.
We use a many-to-one RNN 
for MARKOV and CONCAT.  Given a loss $L$ (we use cross entropy), an initial state $h^{(0)}$, and $\tau$ timesteps each with a sCr measurement $m^{(t)}$, the forward equations with hyperbolic tangent activation are
$$z^{(t)}=Wm^{(t)}+Uh^{(t-1)}+b$$
$$h^{(t)}=\tanh(z^{(t)})$$
$$y^{(\tau )}= Vh^{(\tau)} + c$$

Relative to a many-to-many RNN, the gradient of the hidden state becomes 

$$
\nabla _{h^{(t)}}L = \left\{
        \begin{array}{ll}
            \left(  \dfrac{\partial{y^{}}}{\partial{h^{(t)}}} \right)^{T} \nabla _{y^{}}L 
            = V^{T} \nabla _{y^{}}L & \quad t=\tau \\

            \left( \dfrac{\partial{h^{(t+1)}}}{\partial{h^{(t)}}}\right )^{T} \nabla _{h^{(t+1)}} L
            = 
            
            U^{T}J \nabla _{h^{(t+1)}}L & \quad t \neq  \tau
        \end{array}
    \right.
$$
Where J is the Jacobian of the hyperbolic tangent.

For the NEST input structure, we require an RNN that processes nested, variable-length sequences where each inner sequence produces only a single output. We therefore construct an RNN
for nested sequences by chaining together multiple instances of the many-to-one network (similar to a hierarchical RNN~\cite{chung2016hierarchical}).  We suspect that information might be passed differently from one hospitalization to the next than one measurement to the next, so we introduce a new "rehospitalization" parameter, $R$, that is only between hospitalizations.  Since our data consists of hospitalization sequences that contain measurement sequences, we can index the new equations by $1\leq a \leq A$ where each $a$ has $\tau _ {(a)}$ measurements.  As mentioned in the body of the paper, with $R$, we have the forward equations

 

$$
z^{(t)}_{(a)} = \left\{
        \begin{array}{ll}
            Wm^{(t)} _{(a)}+Rh^{(\tau _ {(a-1)})}_{(a-1)}+r  & \quad t= 1 \\
            
  Wm^{(t)} _{(a)}+Uh^{(t-1)}_{(a)}+b  & \quad t \neq 1 \\
        \end{array}
    \right.
$$

$$h^{(t)} _{(a)}=\tanh(z^{(t)} _{(a)})$$
$$y _{(a)}= Vh^{(\tau _ {(a)})} _{(a)} + c$$

The loss is now summed over hospitalizations from the same patient

$$\mathbb{L}=\sum_{a}L_{(a)} $$

 
  
 


The network is now unrolled in time over measurements and over hospitalizations before backpropagation~\cite{rumelhart1988learning} (automatic differentiation~\cite{rall1981automatic} could also be used here).  Since we use at each hospitalization an RNN with a single output, the hidden state gradient is 



$$
\nabla _{h^{(t)} _{(a)}}L = \left\{
        \begin{array}{ll}

            \left( \dfrac{\partial{h^{(t+1)}_{(a)}}}{\partial{h^{(t)} _{(a)}}}\right )^{T} \nabla _{h^{(t+1)}_{(a)}} L
            = 
            U^{T}J \nabla _{h^{(t+1)}_{(a)}}L & \quad t \neq \tau _{(a)}\\

            
            \left(  \dfrac{\partial{y_{(a)}}}{\partial{h^{(t)} _{(a)}}} \right)^{T} \nabla _{y_{(a)}}L  + \left( \dfrac{\partial{h^{(1)}_{(a+1)}}}{\partial{h^{(t)} _{(a)}}}\right )^{T} \nabla _{h^{(1)}_{(a+1)}} L
            = 
            V^{T} \nabla _{y_{(a)}}L + R^{T}J \nabla _{h^{(1)}_{(a+1)}}L & \quad t =  \tau _{(a)}, a \neq A \\


           \left(  \dfrac{\partial{y_{(a)}}}{\partial{h^{(t)} _{(a)}}} \right)^{T} \nabla _{y_{(a)}}L 
            = V^{T} \nabla _{y_{(a)}}L & \quad t=\tau _{(a)}, a = A \\
        \end{array}
    \right.
$$


The parameter gradients, collected now over time and hospitalizations, are accessible through the node gradients

$$\nabla _{c}L = \sum_{a} \left( \dfrac{\partial{y^{(\tau_{(a)})} _{(a)}}}{\partial{c^{(\tau_{(a)})} _{(a)}}} \right)^{T}\nabla _{y^{(\tau_{(a)})} _{(a)}}L = \sum_{a} \nabla _{y^{(\tau_{(a)})} _{(a)}}L$$

$$\nabla _{V}L
=\sum_{a}  \nabla _{y(\tau_{(a)})}L \left(\dfrac{\partial{y^{(\tau_{(a)})} _{(a)}}}{\partial{V^{(\tau_{(a)})} _{(a)}}}\right)
=\sum_{a} \nabla _{y(\tau_{(a)})}L {h^{(\tau_{(a)})} _{(a)}}^T$$

$$\nabla _{b}L 
= \sum_{a} \sum _{t>1} \left( \dfrac{\partial{h^{(t)} _{(a)}}}{\partial{b^{(t)} _{(a)}}} \right)^{T}\nabla _{h^{(t)} _{(a)}}L = \sum_{a} \sum _{t>1}J\nabla _{h^{(t)} _{(a)}}L$$

$$\nabla _{W}L 
= \sum_{a} \sum _{t}\nabla_{h^{(t)} _{(a)}}L \left(\dfrac{\partial{h^{(t)}_{(a)} }}{\partial{W^{(t)}_{(a)}}} \right)
= \sum_{a} \sum_{t}J \nabla _{h^{(t)} _{(a)}}L {m^{(t)} _{(a)}}^{T}
$$

$$\nabla _{U}L 
= \sum_{a} \sum _{t>1}\nabla_{h^{(t)} _{(a)}}L \left(\dfrac{\partial{h^{(t)} _ {(a)} }}{\partial{U^{(t)}_{(a)}}} \right)
= \sum_{a}  \sum_{t>1}J \nabla _{h^{(t)} _{(a)}}L {h^{(t-1)}_{(a)}}^{T}
$$

$$\nabla _{R}L 
= \sum_{a} \nabla_{h^{(1)} _{(a)}}L \left(\dfrac{\partial{h^{(1)} _ {(a)} }}{\partial{R^{(1)}_{(a)}}} \right)
= \sum_{a}  J \nabla _{h^{(1)} _{(a)}}L {h^{(\tau _{(a-1)})}_{(a-1)}}^{T}
$$

$$\nabla _{r}L 
= \sum_{a} \nabla_{h^{(1)} _{(a)}}L \left(\dfrac{\partial{h^{(1)} _ {(a)} }}{\partial{r^{(1)}_{(a)}}} \right)
= \sum_{a}  J \nabla _{h^{(1)} _{(a)}}L 
$$



            



\newpage
\subsubsection*{Validation Set Distributions}
\begin{figure}[!ht]
\begin{center} 
10 HU
\end{center}
\begin{minipage}{0.33\textwidth}
  \centering
 \includegraphics[width=\linewidth]{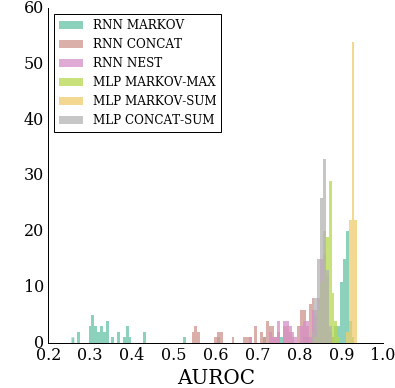}
 \end{minipage} 
 \begin{minipage}{0.33\textwidth}
  \centering
 \includegraphics[width=\linewidth]{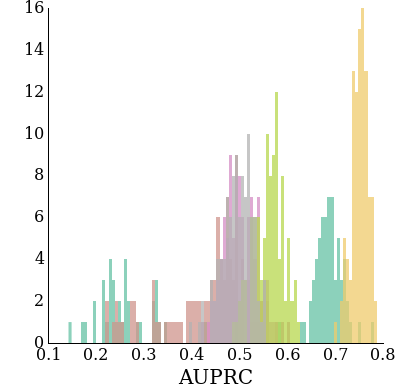}
 \end{minipage} 
  \begin{minipage}{0.33\textwidth}
  \centering
 \includegraphics[width=\linewidth]{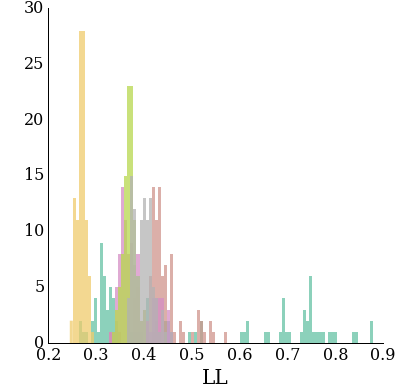}
 \end{minipage} 
\begin{center} 
50 HU
\end{center}
\begin{minipage}{0.33\textwidth}
  \centering
 \includegraphics[width=\linewidth]{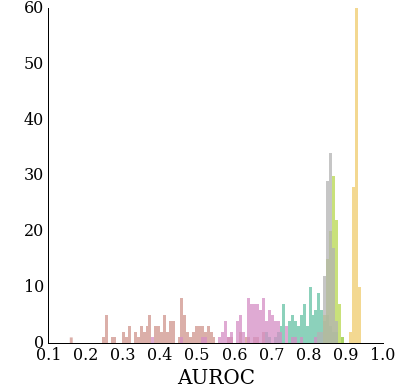}
 \end{minipage} 
 \begin{minipage}{0.33\textwidth}
  \centering
 \includegraphics[width=\linewidth]{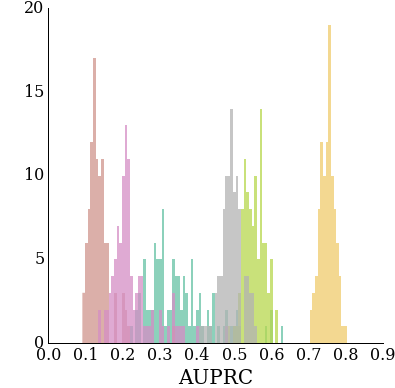}
 \end{minipage} 
  \begin{minipage}{0.33\textwidth}
  \centering
 \includegraphics[width=\linewidth]{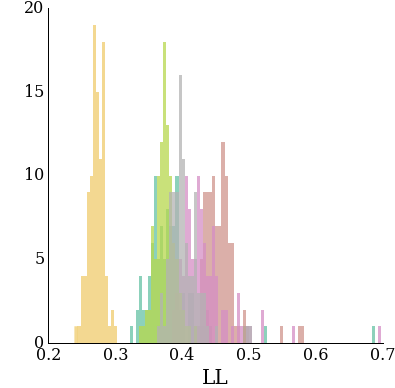}
 \end{minipage} 
 
 \begin{center} 
100 HU
\end{center}
\begin{minipage}{0.33\textwidth}
  \centering
 \includegraphics[width=0.9\linewidth]{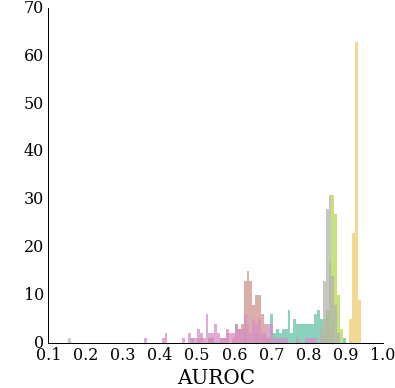}
 \end{minipage} 
 \begin{minipage}{0.33\textwidth}
  \centering
 \includegraphics[width=0.9\linewidth]{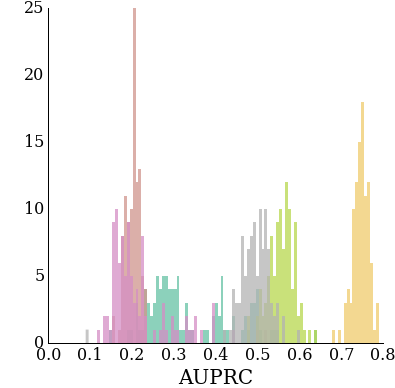}
 \end{minipage} 
  \begin{minipage}{0.33\textwidth}
  \centering
 \includegraphics[width=0.9\linewidth]{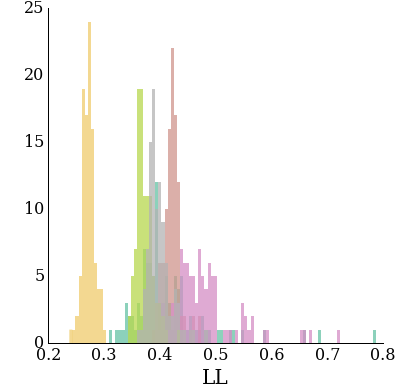}
 \end{minipage} 

  \caption{{\bf Distributions of validation set AUROC, AUPRC, and LL for 10, 50, and 100 hidden unit MLP and RNN processing different inputs structures over 100 trials.} HU = hidden units; LL = log loss; AUPRC = area under PR curve; AUROC = area under ROC curve}  
  \label{fig3}
\end{figure}

\small
\bibliography{b.bib}

\begin{thebibliography}{10}

\bibitem{chertow2005MortLOSCost}
Chertow GM, Burdick E, Honour M, Bonventre JV, Bates DW.
\newblock Acute kidney injury, mortality, length of stay, and costs in
  hospitalized patients [Journal Article].
\newblock J Am Soc Nephrol. 2005;16(11):3365--70.
\newblock Available from: \url{https://www.ncbi.nlm.nih.gov/pubmed/16177006}.

\bibitem{lameire2008prevention}
Lameire N, van Biesen W, Hoste E, Vanholder R.
\newblock The prevention of acute kidney injury an in-depth narrative review:
  Part 2: Drugs in the prevention of acute kidney injury [Journal Article].
\newblock NDT Plus. 2009;2(1):1--10.
\newblock Available from: \url{https://www.ncbi.nlm.nih.gov/pubmed/25949275}.

\bibitem{sutherland2016utilizingAKIADQI}
Sutherland SM, Chawla LS, Kane-Gill SL, Hsu RK, Kramer AA, Goldstein SL, et~al.
\newblock Utilizing electronic health records to predict acute kidney injury
  risk and outcomes: workgroup statements from the 15(th) ADQI Consensus
  Conference [Journal Article].
\newblock Can J Kidney Health Dis. 2016;3:11.
\newblock Available from: \url{https://www.ncbi.nlm.nih.gov/pubmed/26925247}.

\bibitem{rumelhart1986sequential}
Rumelhart DE, Smolensky P, McClelland JL, Hinton G.
\newblock Sequential thought processes in PDP models.
\newblock Parallel distributed processing: explorations in the microstructures
  of cognition. 1986;2:3--57.

\bibitem{koyner2016development}
Koyner JL, Adhikari R, Edelson DP, Churpek MM.
\newblock Development of a Multicenter Ward-Based AKI Prediction Model [Journal
  Article].
\newblock Clin J Am Soc Nephrol. 2016;11(11):1935--1943.
\newblock Available from: \url{https://www.ncbi.nlm.nih.gov/pubmed/27633727}.

\bibitem{choi2016doctor}
Choi E, Bahadori MT, Schuetz A, Stewart WF, Sun J.
\newblock Doctor ai: Predicting clinical events via recurrent neural networks.
\newblock In: Machine Learning for Healthcare Conference; 2016. p. 301--318.

\bibitem{lipton2015learning}
Lipton ZC, Kale DC, Elkan C, Wetzell R.
\newblock Learning to diagnose with LSTM recurrent neural networks.
\newblock arXiv preprint arXiv:151103677. 2015;.

\bibitem{choi2016HF}
Choi E, Schuetz A, Stewart WF, Sun J.
\newblock Using recurrent neural network models for early detection of heart
  failure onset.
\newblock Journal of the American Medical Informatics Association.
  2016;24(2):361--370.

\bibitem{mikolov2013efficient}
Mikolov T, Chen K, Corrado G, Dean J.
\newblock Efficient estimation of word representations in vector space.
\newblock arXiv preprint arXiv:13013781. 2013;.

\bibitem{sordoni2015hierarchical}
Sordoni A, Bengio Y, Vahabi H, Lioma C, Grue~Simonsen J, Nie JY.
\newblock A hierarchical recurrent encoder-decoder for generative context-aware
  query suggestion.
\newblock In: Proceedings of the 24th ACM International on Conference on
  Information and Knowledge Management. ACM; 2015. p. 553--562.

\bibitem{chung2016hierarchical}
Chung J, Ahn S, Bengio Y.
\newblock Hierarchical multiscale recurrent neural networks.
\newblock arXiv preprint arXiv:160901704. 2016;.

\bibitem{ERBPpositionState}
Ad-hoc working~group of E, Fliser D, Laville M, Covic A, Fouque D, Vanholder R,
  et~al.
\newblock A European Renal Best Practice (ERBP) position statement on the
  Kidney Disease Improving Global Outcomes (KDIGO) clinical practice guidelines
  on acute kidney injury: part 1: definitions, conservative management and
  contrast-induced nephropathy [Journal Article].
\newblock Nephrol Dial Transplant. 2012;27(12):4263--72.
\newblock Available from: \url{https://www.ncbi.nlm.nih.gov/pubmed/23045432}.

\bibitem{mckinney2010data}
McKinney W, et~al.
\newblock Data structures for statistical computing in python.
\newblock In: Proceedings of the 9th Python in Science Conference. vol. 445.
  van der Voort S, Millman J; 2010. p. 51--56.

\bibitem{GoodFellowBengio2015deep}
Goodfellow I, Bengio Y, Courville A.
\newblock Deep learning.
\newblock MIT Press; 2016.

\bibitem{walt2011numpy}
Walt Svd, Colbert SC, Varoquaux G.
\newblock The NumPy array: a structure for efficient numerical computation.
\newblock Computing in Science \& Engineering. 2011;13(2):22--30.

\bibitem{duchi2011adaptive}
Duchi J, Hazan E, Singer Y.
\newblock Adaptive subgradient methods for online learning and stochastic
  optimization.
\newblock Journal of Machine Learning Research. 2011;12(Jul):2121--2159.

\bibitem{pedregosa2011scikit}
Pedregosa F, Varoquaux G, Gramfort A, Michel V, Thirion B, Grisel O, et~al.
\newblock Scikit-learn: Machine learning in Python.
\newblock Journal of Machine Learning Research. 2011;12(Oct):2825--2830.

\bibitem{choi2015doctor}
Choi E, Bahadori MT, Sun J.
\newblock Doctor ai: Predicting clinical events via recurrent neural networks.
\newblock arXiv preprint arXiv:151105942. 2015;.

\bibitem{bengio1994learning}
Bengio Y, Simard P, Frasconi P.
\newblock Learning long-term dependencies with gradient descent is difficult.
\newblock IEEE transactions on neural networks. 1994;5(2):157--166.

\bibitem{hochreiter2001gradient}
Hochreiter S, Bengio Y, Frasconi P, Schmidhuber J. Gradient flow in recurrent
  nets: the difficulty of learning long-term dependencies.
\newblock A field guide to dynamical recurrent neural networks. IEEE Press;
  2001.

\bibitem{rumelhart1988learning}
Rumelhart DE, Hinton GE, Williams RJ.
\newblock Learning representations by back-propagating errors.
\newblock Cognitive modeling. 1988;5(3):1.

\bibitem{rall1981automatic}
Rall LB.
\newblock Automatic differentiation: Techniques and applications. 1981;.

\end{thebibliography}



\end{document}